\pdfoutput=1

\documentclass[11pt]{article}

\usepackage[preprint]{acl}

\usepackage{times}
\usepackage{latexsym}

\usepackage[T1]{fontenc}

\usepackage[utf8]{inputenc}

\usepackage[nopatch=footnote]{microtype}

\usepackage{inconsolata}

\usepackage{graphicx}

\usepackage{amsmath}
\usepackage{amssymb}
\usepackage{mathtools}
\usepackage{amsthm}

\usepackage[capitalize,noabbrev]{cleveref}

\usepackage{yhmath}

\usepackage{multirow}
\usepackage{soul}
\usepackage{mathtools}
\usepackage{amsmath,amsfonts,amsthm}
\usepackage{algorithm, algpseudocode}
\usepackage{graphicx}
\usepackage{float}
\usepackage{lipsum}
\usepackage{textcomp}
\usepackage{xcolor}
\usepackage{soul}
\usepackage{placeins}
\usepackage{hhline}
\usepackage{tabularx}
\usepackage{enumitem}
\usepackage{setspace}
\usepackage{booktabs}
\usepackage{adjustbox}
\usepackage[normalem]{ulem}
\usepackage{indentfirst}
\usepackage{titlesec}
\usepackage{wrapfig}
\usepackage{array}
\usepackage{colortbl}
\usepackage{subcaption}
\usepackage{comment} 
\usepackage{tcolorbox}   
\tcbuselibrary{breakable}
\graphicspath{{figures/}}

\theoremstyle{plain}

\theoremstyle{definition}

\theoremstyle{remark}

\usepackage{titlesec}

\titlespacing*{\section}{0pt}{0.1\baselineskip}{0.1\baselineskip}

\titlespacing*{\subsection}{0pt}{0.1\baselineskip}{0.1\baselineskip}

\titlespacing*{\subsubsection}{0pt}{0.1\baselineskip}{0.1\baselineskip}

\title{Learning User Interests via Reasoning and Distillation for Cross-Domain News Recommendation}

\author{
  \textbf{Mengdan Zhu}\textsuperscript{1,2}\thanks{Work done during internship at Microsoft.} \quad
  \textbf{Yufan Zhao}\textsuperscript{1} \quad
  \textbf{Tao Di}\textsuperscript{1} \quad
  \textbf{Yulan Yan}\textsuperscript{1} \quad
  \textbf{Liang Zhao}\textsuperscript{2}
  \\
  \\
  \textsuperscript{1}Microsoft \quad
  \textsuperscript{2}Emory University
  \\
  \\ 
  \scalebox{0.95}{\ttfamily \{mengdan.zhu, liang.zhao\}@emory.edu}  \\
   \scalebox{0.95}{\ttfamily{\{Yufan.Zhao, Tao.Di, yulanyan\}@microsoft.com}}
}

\begin{document}
\maketitle

\begin{abstract}
News recommendation plays a critical role in online news platforms by helping users discover relevant content. Cross-domain news recommendation further requires inferring user's underlying information needs from heterogeneous signals that often extend beyond direct news
consumption. A key challenge lies in moving beyond surface-level behaviors to capture deeper, reusable user interests while maintaining scalability in large-scale production systems. In this paper, we present a reinforcement learning framework that trains large language models to generate high-quality lists of interest-driven news search queries from cross-domain user signals. We formulate query-list generation as a policy optimization problem and employ GRPO with multiple reward signals. We systematically study two compute dimensions: inference-time sampling and model capacity, and empirically observe consistent improvements with increased compute that exhibit scaling-like behavior. Finally, we perform on-policy distillation to transfer the learned policy from a large, compute-intensive teacher to a compact student model suitable for scalable deployment. Extensive offline experiments, ablation studies and large-scale online A/B tests in a production news recommendation system demonstrate consistent gains in both interest modeling quality and downstream recommendation performance.
\end{abstract}

\section{Introduction}
\label{sec:introduction}
Personalized news recommendation plays a central role in modern online news platforms \cite{google_news, lavie2010user} by helping users efficiently discover relevant content under severe information overload.  Large-scale news services stream tens of thousands of articles from diverse sources every day, far exceeding what users can manually consume. As a result, effective recommendation systems are critical for matching user interests, improving engagement, and enabling scalable content discovery in online news ecosystems \cite{ijntema2010ontology}.

Recent advances in large language models (LLMs) have further expanded the design space of recommendation systems \cite{Wu2024LLM4RecSurvey}. Owing to their strong reasoning ability \cite{wei2023chainofthoughtpromptingelicitsreasoning}, rich world knowledge, and semantic generalization capacity \cite{shehmir2025llm4rec}, LLMs offer new opportunities to move beyond traditional collaborative filtering and embedding-based approaches that primarily rely on historical interaction similarity. In contrast, LLMs can reason over heterogeneous user signals and generate semantically meaningful representations that capture higher-level interests, enabling more flexible, diverse, and potentially long-term-oriented recommendation strategies \cite{zhang2025collmintegratingcollaborativeembeddings}.

The limitations of the traditional embedding-based cross-domain recommendation methods are that they are hard to learn reasonable and personalized embeddings for all users~\cite{zhu2021learning}, and collaborative filtering methods often fail to capture deeper sequential signals~\cite{he2017neural}. However, cross-domain recommendation demands deeper user understanding beyond shallow behavior patterns. Our approach that leverages LLMs to reason over heterogeneous user signals could effectively addresses these limitations.

We reformulate cross-domain interest discovery as a query-list generation problem for news recommendation. Given heterogeneous user signals from multiple user interaction domains, we train an instruction-tuned LLM to generate lists of news search queries that represent reusable user interests. The model is optimized using Group Relative Policy Optimization (GRPO) with five reward signals, including retrieval alignment, interest coverage, query specificity, intra-list diversity, and structural validity.

We systematically study the effect of additional compute along two dimensions: \textbf{inference-time sampling} and \textbf{model capacity}. Increasing either dimension consistently improves interest quality and downstream retrieval performance, exhibiting clear scaling-like behavior. To make the approach practical for online serving, we further apply on-policy distillation. This transfers the learned policy from a large teacher model to a compact student model with low latency and high throughput.

This work makes the following contributions:
\begin{itemize}[leftmargin=*, itemsep=2pt, topsep=4pt]
   \item \textbf{Scaling properties of LLMs for interest modeling under noisy signals.}
    We show that, when trained with multiple product-aligned reward models, our models can effectively infer reusable user interests from heterogeneous and noisy cross-domain signals. Moreover, both increasing model capacity (space scaling) and inference-time sampling (time scaling) lead to consistent, scaling-law–like improvements in interest quality and downstream retrieval performance.

    \item \textbf{On-policy distillation for scalable interest generation.}
    We propose an on-policy distillation framework that transfers the learned, compute-intensive LLM policy into a compact student model that performs lightweight reasoning to generate query lists. The distilled model preserves most of the interest modeling and retrieval gains while enabling low-latency, high-throughput serving in production systems.

    \item \textbf{Validation on offline evaluations and online A/B tests.}
    We demonstrate the effectiveness of the proposed approach through comprehensive offline evaluation and controlled online A/B tests, showing consistent improvements in interest modeling quality and downstream recommendation performance.

\end{itemize}
To the best of our knowledge, this work represents a pioneering \textbf{deployment} of a reasoning-driven reinforcement learning framework for user interest modeling in a large-scale commercial news recommendation system.

\section{Related Work}
\label{sec:related work}

\noindent \textbf{User Interest Modeling and Sequential Recommendation.}  
User interest modeling typically learn user preferences from historical interactions such as clicks, queries, or browsing histories. The field has transitioned from traditional collaborative filtering to deep sequential modeling. Early approaches utilized RNN architectures \cite{okura2017embedding, hidasi2015session} to capture temporal patterns. The adoption of self-attention mechanisms marked a significant shift, enabling models to capture long-range item dependencies \cite{kang2018self, sun2019bert4rec}. In the context of news recommendation, models like NRHUB \cite{wu2019neural} and hierarchical frameworks \cite{Qi2021HieRec} were developed to capture multi-level user interests. For large-scale industrial graphs, PinSage \cite{pal2020pinnersage} demonstrated the power of random-walk graph convolutions. More recently, the HSTU \cite{zhai2024actions} framework, have scaled these concepts to trillion-parameter generative transducers for industry-scale recommendations.

\noindent \textbf{LLMs for RecSys.}  
LLMs have recently been explored for recommendation systems (RecSys) \cite{Wu2024LLM4RecSurvey}. LLMs for RecSys can be categorized into \emph{non-generative} and \emph{generative} paradigms \cite{wang2024towards}. Non-generative approaches keep the conventional "score-then-rank" structure but use LLMs for feature engineering \cite{xi2024towards}, or representation learning \cite{hou2023learning}. Generative approaches instead generate recommendations directly. This paradigm typically involves prompting, which utilizes prompt engineering to leverage the LLM’s zero-shot or few-shot capabilities \cite{hou2024large}, and fine-tuning, where models are fine-tuned on recommendation datasets to align their generative outputs with user preferences \cite{lin2024data}. Unlike traditional fine-tuning, our work explores using GRPO to optimize the reasoning trajectories for a query-list deep user interest generation in recommendation systems. 

\section{System Overview}
\label{sec:system-overview}

\begin{figure*}[t]
    \centering
    \includegraphics[width=\textwidth]{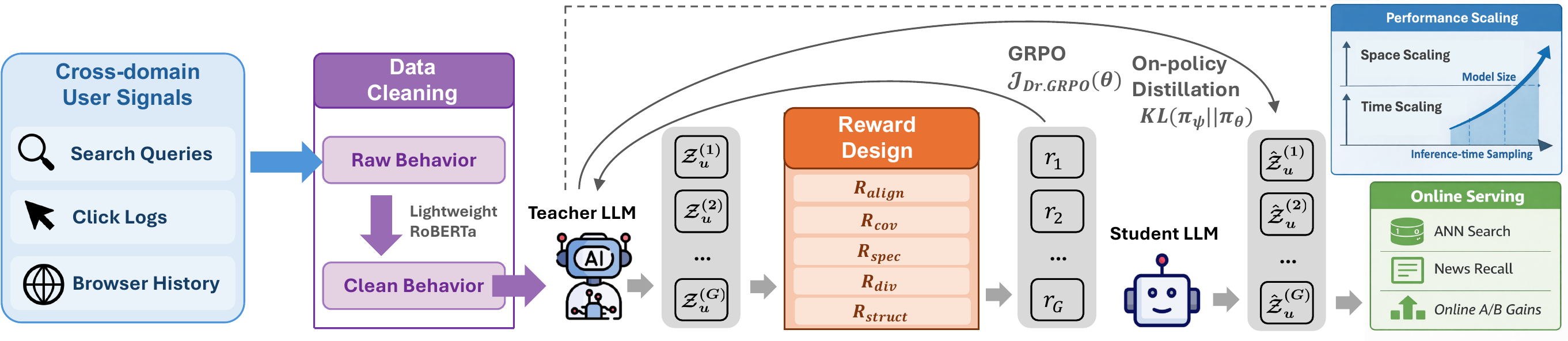}
    \caption{Overview of the Reasoning-driven User Interest Generation for News Search Queries (Teacher Model) and On-policy Distillation for Online Serving (Student Model) in the large-scale RecSys.}
    \label{fig:main}
\end{figure*}

Our system operates on multiple streams of user behavioral data collected across search logs, click logs, browsing interactions, and news engagements. Each user is represented using anonymized hashed identifiers to ensure privacy. These heterogeneous data sources capture a broad spectrum of user behaviors and thus provide strong signals for generating high-quality lists of interest-driven news search queries for downstream retrieval. 

Our framework consists of three stages as illustrated in Fig.~\ref{fig:main}:
\begin{itemize}[leftmargin=*, itemsep=2pt, topsep=4pt]
    \item Data Cleaning: Raw behavioral logs are inherently noisy, mixed with malformed or irrelevant search queries (e.g., random strings, phone numbers, navigational inputs), and behaviors that are less correlated with users' underlying interests. Consequently, it is essential to filter out these irrelevant or non-informative signals prior to the user interest modeling.
    \item User Interest Generation: Cleaned behaviors are fed into a strong teacher LLM, which generates a list of interest-driven news search queries under a reinforcement learning framework with multiple reward signals.
    \item On-Policy Distillation: In practice, one key challenge for LLM-based recommendation is compute-intensive. While the teacher LLM can generate high-quality interests, its inference is computationally heavy, incurring substantial serving cost that are incompatible with large-scale online recommendation. To bridge this deployment gap, we distill the teacher policy into a lightweight student model for production. In online serving, the student generates a list of interest queries, which is then used to retrieve candidate news articles via ANN-based retrieval.
\end{itemize}


\section{Method}
\label{sec:method}

Our goal is to infer a list of news search queries that represent reusable deep user interests from noisy cross-domain behavioral logs in a large-scale commercial news recommendation system. We begin by formalizing the problem (Section~\ref{subsec:problem}), then describe the noise cleaning pipeline (Section~\ref{subsec:reasoning}), followed by our interest-generation model trained with multi-reward reinforcement learning (Section~\ref{subsec:rl}). Finally, we present our on-policy distillation procedure that distills the teacher’s reasoning capability to an efficient student model for online serving (Section~\ref{subsec:onpolicy}).

\subsection{Problem Formulation}
\label{subsec:problem}

Let $u$ denote a user and $\mathcal{B}_u = \{b_1, b_2, \dots, b_{t}\}_{u}$ denote the multiset of raw behavioral signals observed for that user across various platforms. 

Our objective is to infer a list of user interests as news search queries:
\vspace{-2mm}
\begin{equation}
    \mathcal{Z}_u = \{ z_1, z_2, \dots, z_n\}_u,
\end{equation}
where each $z_i$ is a short natural-language phrase describing a user interest.

We model interest generation as a policy optimization problem:
\vspace{-2mm}
\begin{equation}
    \pi_\theta(\mathcal{Z}_u \mid \mathcal{B}_u),
\end{equation}
where $\pi_{\theta}$ outputs a list of $n$ queries given cross-domain signals. 

We adopt a three-stage training strategy: (1) filter out noisy user signals, (2) optimize a teacher policy with GRPO to directly align generated interest queries with retrieval utility and desired interest properties, and (3) distill the learned policy into a student model that is efficient for online serving.



\subsection{Noise Cleaning}
\label{subsec:reasoning}

Real-world raw behavioral logs often contain substantial noisy and irrelevant information that does not reflect a user's underlying interest. To mitigate the impact of this noise on the downstream reasoning process, we employ a lightweight binary classifier $f_{\phi}$. We obtain supervision by annotating a subset of behavioral signals using GPT5 with a designed prompt (see Appendix~\ref{app:prompt_noise}), and train a RoBERTa-based classifier to distinguish informative behaviors from irrelevant noise. For each raw behavioral signal $b_t$, the classifier assigns a label $y_t \in \{0, 1\}$, where $y_t = 1$ indicates a signal that should be retained for interest modeling, and $y_t = 0$ indicates noise to be discarded. The refined behavioral set $\tilde{\mathcal{B}}_u$ is formally defined as:
\begin{equation}
\tilde{\mathcal{B}}_u = \{ b_t \mid f_{\phi}(b_t) = 1 \}_u.
\end{equation}

\subsection{Interest Generation with Multi-Reward Reinforcement Learning}
\label{subsec:rl}

To derive high-quality interest queries from cleaned cross-domain behaviors $\tilde{\mathcal{B}}_u$ without ground-truth labels $\mathcal{Z}^*_u$, we formulate the interest discovery process as a policy optimization problem with a composite reward that reflects downstream retrievability and desired interest properties.

\paragraph{Policy Formulation.}  
Inspired by the self-evolution capabilities observed in DeepSeek-R1-Zero~\cite{guo2025deepseek}, we adopt Dr.GRPO~\cite{liu2025understandingr1zeroliketrainingcritical}, an improved variant of GRPO, to guide the policy towards maximizing our multi-dimensional rubric objectives, using the group average as a baseline.


Formally, given a user $u$ with cleaned behaviors $\tilde{\mathcal{B}}_u$, the policy $\pi_\theta$ generates a group of $G$ candidate interest lists 
\begin{equation}
\{\mathcal{Z}_u^{(1)}, \mathcal{Z}_u^{(2)}, \dots, \mathcal{Z}_u^{(G)}\},
\end{equation}
where each $\mathcal{Z}_u^{(g)} = \{ z_{u,1}^{(g)}, \dots, z_{u,n}^{(g)} \}$ is a list of $n$ interest queries sampled from the old policy $\pi_{\theta_{\text{old}}}$.

Each list $\mathcal{Z}_u^{(g)}$ receives a scalar reward $r_g = \mathcal{R}(u, \mathcal{Z}_u^{(g)})$, and the objective maximizes the expected return over the sampled group.

The objective function is defined to maximize the expected return:

\vspace{-6mm}
\begin{equation}
\begin{aligned}
\mathcal{J}_{\text{Dr.GRPO}}(\theta)
&= \mathbb{E} \Bigg[
\frac{1}{G}
\sum_{g=1}^{G}
\ell_g(\theta)
\Bigg]
- \beta D_{KL}(\pi_\theta \,\|\, \pi_{\text{ref}}),
\end{aligned}
\end{equation}

\noindent where
\begin{equation}
\ell_g(\theta)
=
\min\Big(
\rho_g \hat{A}_g,\;
\operatorname{clip}(\rho_g,1-\epsilon,1+\epsilon)\hat{A}_g
\Big),
\end{equation}

\noindent the probability ratio for the $g$-th sampled interest list is defined as
\vspace{-1mm}
\begin{equation}
\rho_g = 
\frac{\pi_\theta(\mathcal{Z}_u^{(g)} \mid \tilde{\mathcal{B}}_u)}
{\pi_{\theta_{\text{old}}}(\mathcal{Z}_u^{(g)} \mid \tilde{\mathcal{B}}_u)},
\end{equation}
with $\mathcal{Z}_u^{(g)}$ denoting the $g$-th sampled interest query list for user $u$.

The advantage term is computed using a group-relative baseline:
\vspace{-3mm}
\begin{equation}
\hat{A}_g = r_g - \frac{1}{G} \sum_{j=1}^{G} r_j.
\end{equation}

\paragraph{Reward Design.}  
We decompose the quality of the generated interest list $\mathcal{Z}_u$ into five complementary criteria:
\vspace{-2mm}
\begin{equation}
    \mathcal{R}(u, \mathcal{Z}_u)=\sum_{m=1}^5 \lambda_{m}R_m(u, \mathcal{Z}_u),
\end{equation}
where:
\begin{itemize}[leftmargin=*, itemsep=2pt, topsep=4pt]
    \item \textbf{Retrieval Alignment ($R_{\text{align}}$)}: Ensures that each generated query can successfully retrieve relevant articles from our internal ANN-based index.
    \item \textbf{Interest Coverage ($R_{\text{cov}}$)}: Encourages the generated query set to collectively reflect the diverse latent interests inferred from user behaviors.
    \item \textbf{Query Specificity ($R_{\text{spec}}$)}: Encourages content-rich and non-generic queries that better capture personalized user intent.
    \item \textbf{Intra-list Diversity ($R_{\text{div}}$)}: Promotes semantic non-redundancy among generated queries within the same list.
    \item \textbf{Structural Validity ($R_{\text{struct}}$)}: A binary reward verifying whether the output satisfies predefined structural constraints.
\end{itemize}


In practice, $R_{\text{align}}$ is computed by retrieving, for each query, its top-$K$ articles ($K=10$) from our internal ANN-based news index and averaging the embedding similarities as a soft alignment score. $R_{\text{div}}$ is calculated as the average pairwise cosine distance between query embeddings obtained from Qwen3-Embedding-8B \cite{qwen3embedding}. Both $R_{\text{cov}}$ and $R_{\text{spec}}$ are implemented using Rubrics as Rewards \cite{gunjal2025rubrics}, where an LLM evaluator assesses interest coverage and query specificity based on the prompt template provided in Appendix~\ref{app:prompt_reward}. $R_{\text{struct}}$ is a binary rule-based score verifying adherence to the required output format. All $R_m$ are normalized before computing the overall reward $\mathcal{R}$.

\subsection{On-Policy Distillation of Interest Models}
\label{subsec:onpolicy}

\noindent While the GRPO-trained teacher produces high-quality interest lists, direct online serving is often infeasible due to latency and high compute cost. We therefore transfer the teacher’s policy into a compact student model via \emph{on-policy distillation}~\cite{lu2025onpolicy}, which combines (i) the exploration coverage of student rollouts with (ii) dense token-level supervision from the teacher.

We formulate the distillation objective as minimizing the expected reverse Kullback-Leibler (KL) Divergence between the student $\pi_\psi$ and teacher $\pi_\theta$ distributions:

{\small
\vspace{-4mm}
\begin{equation}
\mathcal{L}_{distill}(\psi)=\mathbb{E}_{y_{<t}\sim\pi_{\psi}}[D_{KL}(\pi_{\psi}(\cdot|\tilde{\mathcal{B}}_{u},y_{<t})||\pi_{\theta}(\cdot|\tilde{\mathcal{B}}_{u},y_{<t}))].
\end{equation}}

At each decoding step $t$, we evaluate the divergence between the student and teacher distributions conditioned on the student's generated prefix $y_{<t}$ and encourage the student to approximate the teacher’s distribution at that prefix.

\section{Experiments}
\label{sec:experiment}
\subsection{Experimental Setup}
\paragraph{Datasets. }We conduct experiments on real-world user interaction logs collected from an internal, large-scale production news recommendation system. The dataset includes cross-domain signals from web browsing, search queries, and news recommendation logs. For each user, we retain up to the most recent 50 interaction events in each domain, and a subset of users is sampled for experiments. Data from November 2025 is used for training and validation (with 10\% held out for validation), while data from the first week of December 2025 is reserved for evaluation based on downstream news retrieval and user engagement. The statistics of the processed dataset are summarized in Table~\ref{tab:dataset_statistics}.

\paragraph{Evaluation Metrics.}
We evaluate performance using standard retrieval and ranking metrics, including Recall@K, NDCG@K, and MRR, where $K \in \{5, 10\}$. All metrics are computed on the held-out test set and averaged across users.

\paragraph{Baselines.}To evaluate the performance of the proposed method, we compare it with several representative baselines. 
GRU~\cite{okura2017embedding} is an early neural news recommendation model that employs an autoencoder with GRU to learn news and user representations. 
SASRec~\cite{kang2018self} introduces self-attention to model sequential user behaviors. 
NRHUB~\cite{wu2019neural} models user interests by aggregating heterogeneous behavioral signals from multiple interaction types. 
PinSage~\cite{pal2020pinnersage} represents user interests by clustering interaction embeddings using a hierarchical (Ward) clustering strategy. 
More recently, HSTU~\cite{zhai2024actions} adopts a Transformer-based architecture to capture fine-grained sequential user behaviors.

\paragraph{Implementation Details.}
We use Qwen2.5-32B-Instruct as the teacher model and Qwen2.5-0.5B-Instruct as the student model. The teacher model is trained with GRPO to generate lists of interest-driven search queries from heterogeneous cross-domain user signals. The learned policy is then distilled on-policy into the student model for low-latency serving (see Appendix~\ref{app:efficiency}). All experiments are conducted on a cluster of 128 NVIDIA A100 GPUs. During evaluation, each generated interest query is used to retrieve candidate news articles via ANN-based retrieval, and the top-$K$ retrieved items are aggregated for downstream evaluation. For all baseline methods, we follow their original implementations and recommended settings, and conduct extensive hyperparameter tuning to ensure a fair comparison.

\begin{table}[t]
\centering
\setlength{\tabcolsep}{3pt}
\renewcommand{\arraystretch}{0.9}
\begin{tabular}{lc}
\toprule
\textbf{Statistic} & \textbf{Overall} \\
\midrule
\# Users & 500,000 \\
\# News & 1,214,552 \\
Avg. browsing events per user & 37 \\
Avg. search queries per user & 13 \\
Avg. news clicks per user & 39 \\
\bottomrule
\end{tabular}
\caption{Statistics of the processed dataset.}
\label{tab:dataset_statistics}
\end{table}

\begin{table*}[t]
\centering
\setlength{\tabcolsep}{4pt}
\renewcommand{\arraystretch}{0.85}
\begin{tabular}{lccccc}
\toprule
\textbf{Method} 
& \textbf{Recall@5} 
& \textbf{Recall@10} 
& \textbf{NDCG@5} 
& \textbf{NDCG@10} 
& \textbf{MRR} \\
\midrule
GRU~\cite{okura2017embedding}      & 0.144 & 0.189 & 0.129 & 0.159 & 0.117 \\
SASRec~\cite{kang2018self}        & 0.158 & 0.206 & 0.141 & 0.173 & 0.129 \\
NRHUB~\cite{wu2019neural}         & 0.162 & 0.212 & 0.145 & 0.178 & 0.133 \\
PinSage~\cite{pal2020pinnersage}  & 0.168 & 0.220 & 0.149 & 0.184 & 0.138 \\
HSTU~\cite{zhai2024actions}       & 0.216 & 0.250 & 0.194 & 0.228 & 0.184 \\
\midrule
\textbf{Ours (RL-trained, Qwen2.5-32B)} & \textbf{0.257} & \textbf{0.271} & \textbf{0.209} & \textbf{0.241} & \textbf{0.199} \\
\textbf{Ours (Distilled, Qwen2.5-0.5B)} & 0.233 & 0.261 & 0.196 & 0.235 & 0.180 \\
\bottomrule
\end{tabular}
\caption{Evaluations of methods on the offline dataset.}
\label{tab:main_results}
\end{table*}

\label{subsec:setup}
\subsection{Main Results}
Table~\ref{tab:main_results} reports the main results on the test set. Overall, our approach consistently outperforms all baseline methods across all evaluation metrics. Among the baselines, HSTU~\cite{zhai2024actions} achieves the strongest performance, highlighting the advantage of large-capacity sequential models in capturing complex user behavior patterns.

Building on this strong baseline, both our RL-trained teacher model and the distilled student model further improve performance, indicating that explicitly learning reusable and abstract user interests from heterogeneous cross-domain signals provides complementary benefits beyond sequence modeling alone. The RL-trained teacher model based on Qwen2.5-32B achieves the best results across all metrics, demonstrating the effectiveness of reinforcement learning for interest-driven query generation. Notably, the distilled Qwen2.5-0.5B model preserves most of the performance gains while substantially reducing model size, suggesting that on-policy distillation can effectively transfer the learned interest modeling capability to a compact model suitable for efficient deployment.

\begin{figure}[t]
    \centering
    \includegraphics[width=0.9\columnwidth]{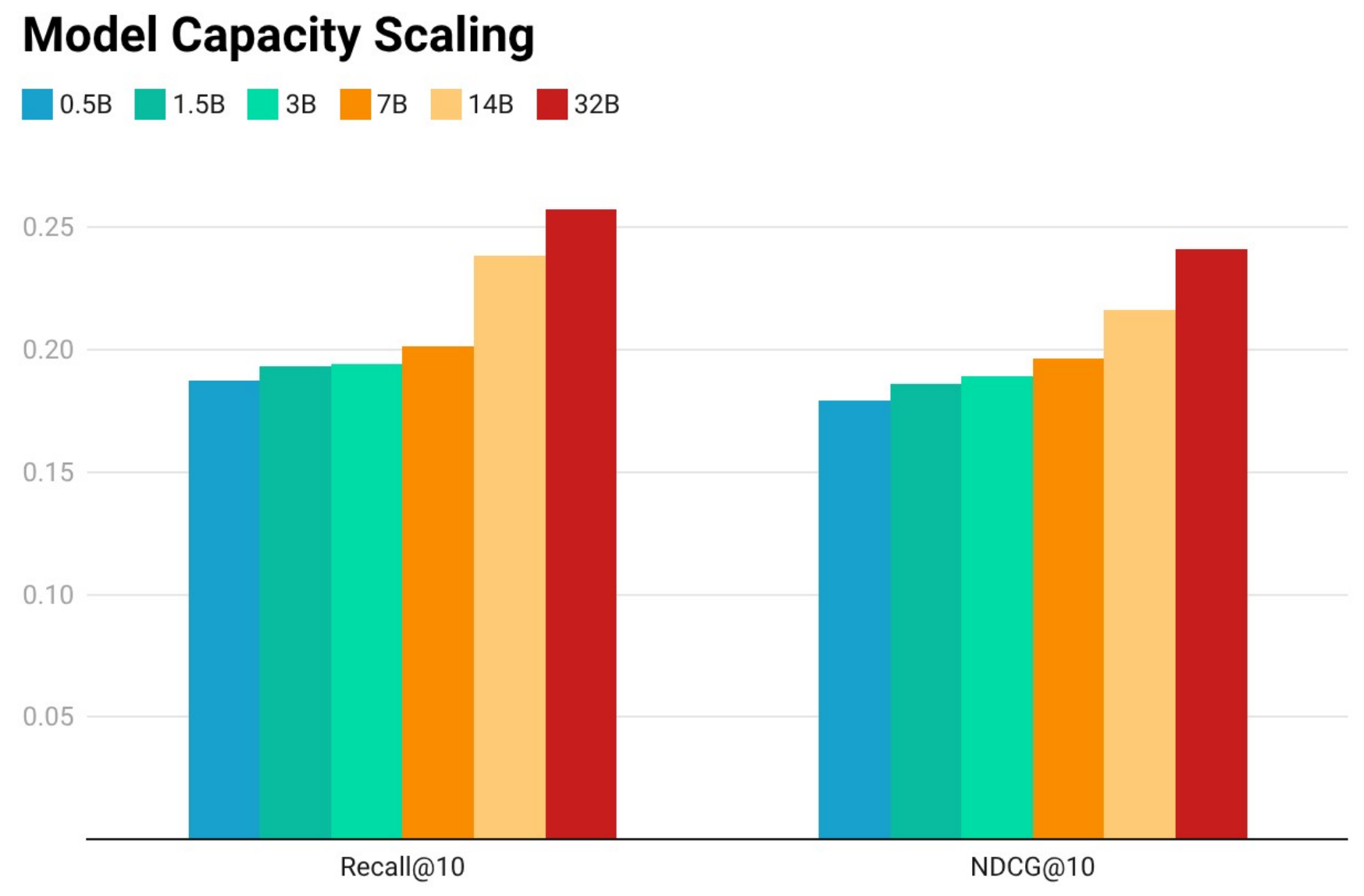}
    \caption{Model capacity scaling for interest generation. All models belong to the Qwen2.5 family and are trained with identical objectives and inference settings. Increasing parameter size from 0.5B to 32B consistently improves Recall@10 and NDCG@10, demonstrating scaling-like behavior in deep user interest modeling.
}
    \label{fig:scaling}
\end{figure}

\label{subsec:setup}
\subsection{Ablation Study}
\begin{figure}[t]
    \centering
    \includegraphics[width=0.85\columnwidth]{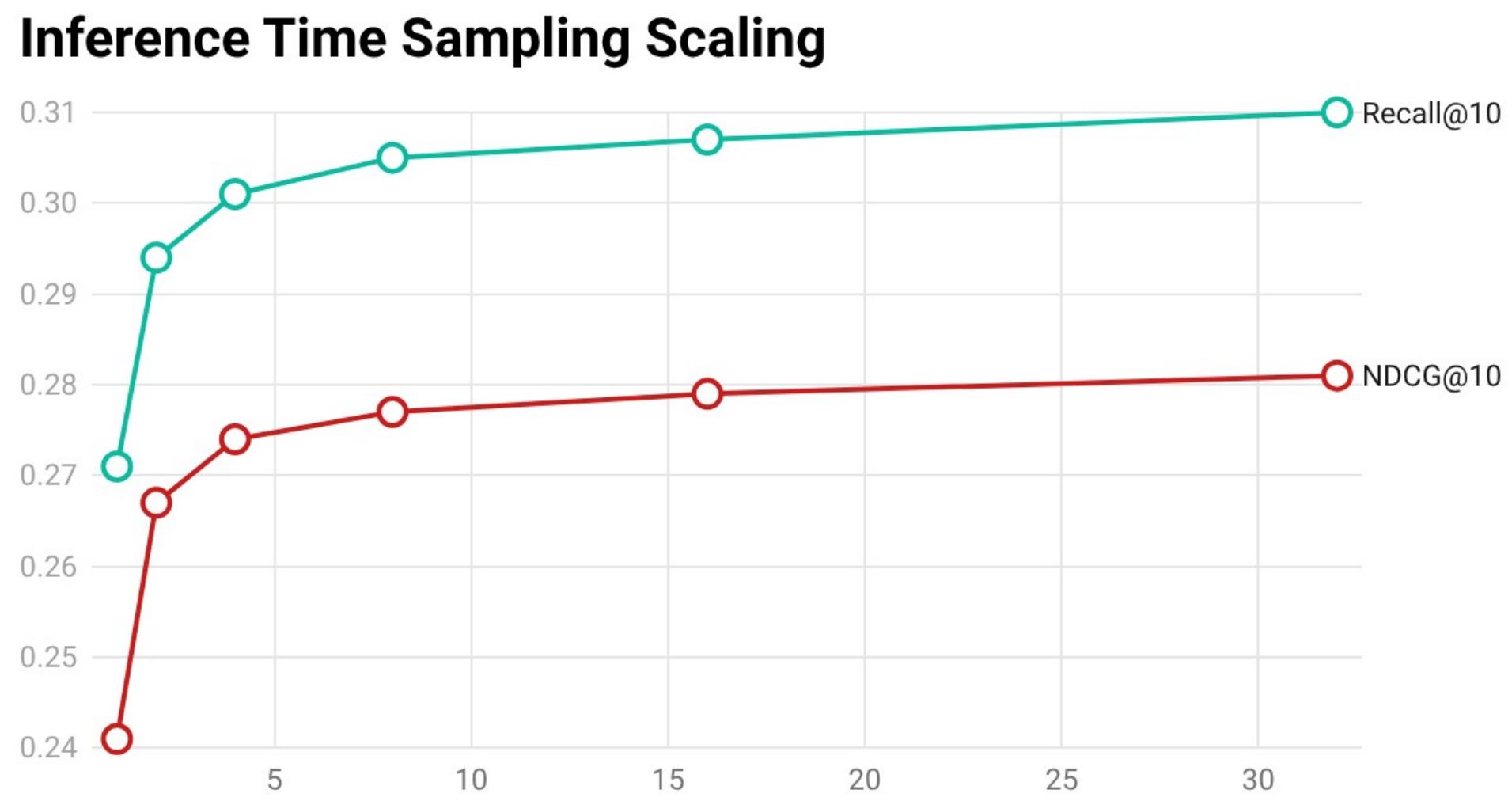}
    \caption{Inference-time sampling scaling with best-of-$N$ selection. Increasing the number of sampled candidate interest lists ($N \in \{1,2,4,8,16,32\}$) consistently improves Recall@10 and NDCG@10, exhibiting diminishing returns at larger $N$.}
    \label{fig:test_time_scaling}
\end{figure}

\begin{table}[t]
\centering
\setlength{\tabcolsep}{3pt}
\renewcommand{\arraystretch}{0.9}
\small
\begin{tabular}{lcc}
\toprule
\textbf{Distillation Method} & \textbf{Recall@10} & \textbf{NDCG@10} \\
\midrule
Teacher (Qwen2.5-32B) & 0.271 & 0.241 \\
\midrule
Supervised Distillation & 0.245 & 0.207 \\
On-policy Distillation & \textbf{0.261} & \textbf{0.235} \\
\midrule
No Distillation & 0.187 & 0.179 \\
\bottomrule
\end{tabular}
\caption{Comparison of different distillation strategies.}
\label{tab:distillation_ablation}
\end{table}

\noindent \textbf{Model Size Scaling.}
We analyze the impact of model capacity by varying the size of the interest generation model while keeping the training procedure and inference settings fixed. As shown in Fig.~\ref{fig:scaling}, increasing model size consistently improves retrieval metrics and interest quality, indicating that larger models enable stronger abstraction and reasoning over cross-domain user behaviors, resulting in scaling-like performance gains.

\noindent \textbf{Inference-time Sampling Scaling.}
We analyze time scaling by increasing the number of sampled candidate interest lists at inference time. As shown in Fig.~\ref{fig:test_time_scaling}, for each user we generate $N \in \{1,2,4,8,16,32\}$ lists and select the final output via a \emph{best-of-$N$} strategy based on the composite reward $\mathcal{R}$. Increasing $N$ consistently improves downstream retrieval performance with diminishing returns, indicating additional headroom from inference-time compute in offline evaluation. In production, we use $N=1$ due to latency constraints.

\noindent \textbf{On-Policy Distillation.}
We evaluate on-policy distillation by comparing it with supervised distillation and training without teacher guidance. As shown in Table~\ref{tab:distillation_ablation}, on-policy distillation consistently outperforms supervised distillation and closes much of the gap to the teacher model, indicating more effective supervision under low-latency serving constraints.

\subsection{Online Experiment}
\label{subsec:online}

\begin{table}[t]
\centering
\begin{tabular}{lcc}
\toprule
\textbf{Metric} & \textbf{Lift (\%)} & \textbf{p-value} \\
\midrule
DAU & +0.95\% & 0.012 \\
CTR & +0.22\% & 0.031 \\
Cold User DAU & +4.38\% & 0.004 \\
\bottomrule
\end{tabular}
\caption{Online A/B experiment results over a 7-day flight. All lifts are statistically significant.}
\label{tab:online_results}
\end{table}

We further evaluate the proposed approach through an online experiment in a large-scale production news recommendation system. The method is deployed as an additional recall path alongside existing retrieval components, without replacing any existing modules. We conduct a 7-day online A/B flight experiment, comparing the proposed approach against the production baseline under identical traffic allocation.

As shown in Table~\ref{tab:online_results}, the proposed method leads to consistent improvements in DAU and CTR during the online flight. Notably, we observe a substantially larger gain in cold-user DAU. This suggests that leveraging heterogeneous cross-domain signals enables the model to better infer user interests even when direct news interaction history is limited, resulting in more personalized and relevant recommendations for cold users.

\section{Conclusion}
\label{sec:conclusion}

In this paper, we presented a reasoning-driven reinforcement learning framework for cross-domain user interest modeling in large-scale news recommendation systems. Extensive offline evaluation, ablation studies, and large-scale online A/B tests further validate the effectiveness of our method. This work represents an important step toward deployable reasoning-centric reinforcement learning for recommendation systems, and opens new directions for combining LLM reasoning, scalable RL, and industrial recommendation pipelines.


\bibliography{references}

\clearpage
\appendix
\section{Prompt for the Rubrics-as-Rewards.}
\label{app:prompt_reward}

\begin{tcolorbox}[
breakable,
  colback=gray!5!white,
  colframe=gray!50!black,
  title={Prompt for Rewards $R_{\text{spec}}$ and $R_{\text{cov}}$}]

\textbf{System Prompt}

You are a strict evaluator for a user interest query generation system in a large-scale news recommendation setting.
Score the generated query list using the rubrics below. Be consistent and conservative. Do not reward generic or vague queries.
Return JSON only.

\medskip
\textbf{User Prompt}

Input: Cleaned Cross-domain User Behaviors
\{USER\_BEHAVIOR\_CONTEXT\}

Model Output: Generated Interest Query List
\{GENERATED\_QUERY\_LIST\}

\medskip
\textbf{Rubrics}

\paragraph{(1) Query Specificity ($R_{\text{spec}}$)}
Evaluate each query individually and then average.

For each query $z_i$, assign $w_i \in \{0,1\}$:
\begin{itemize}
  \item \textbf{1} if $z_i$ is specific and content-rich (e.g., includes concrete topics/events/entities, clear constraints, or an unambiguous intent).
  \item \textbf{0} if $z_i$ is generic, overly broad, ambiguous, or mainly a platform/common category term (e.g., ``news'', ``sports'', ``technology'', ``YouTube'', ``Google'').
\end{itemize}

Compute:
\[
R_{\text{spec}}=\frac{1}{n}\sum_{i=1}^{n} w_i.
\]

\paragraph{(2) Interest Coverage ($R_{\text{cov}}$)}
First, extract a set of distinct major interest themes from the user behaviors, then measure whether the query list covers them.

\textbf{Step A (Theme extraction):} Extract a set of distinct \emph{major} interest themes from the behaviors. Use as many themes as necessary to summarize the behaviors without being overly granular (i.e., do not split into near-duplicate subtopics). Each theme should be short (a few words) and represent a coherent interest area.

\textbf{Step B (Theme coverage):} For each theme $t_j$, assign $c_j \in \{0,1\}$:
\begin{itemize}
  \item \textbf{1} if at least one query in the list clearly targets theme $t_j$.
  \item \textbf{0} if no query clearly covers $t_j$ (weak or indirect matches should be scored 0).
\end{itemize}

Let $K$ be the number of extracted themes. Compute:
\[
R_{\text{cov}}=\frac{1}{K}\sum_{j=1}^{K} c_j.
\]

\medskip
\textbf{Output Format}

Return a JSON object \emph{only}:

\begin{verbatim}
{
  "R_spec": <float between 0 and 1>,
  "R_cov": <float between 0 and 1>,
  "spec_per_query": [0 or 1, ...],
  "themes": ["...", "...", "..."],
  "covered": [0 or 1, ...]
}
\end{verbatim}

\end{tcolorbox}

\section{Prompt for Noise Cleaning}
\label{app:prompt_noise}

\begin{tcolorbox}[
breakable,
  colback=gray!5!white,
  colframe=gray!50!black,
  title={Prompt for Behavioral Noise Filtering}]

\textbf{System Prompt}

You are a strict classifier for filtering user behavioral signals in a large-scale news recommendation system.

Your task is to determine whether a single user behavior reflects a meaningful informational interest that can contribute to long-term news interest modeling.

Be conservative. 
Platform navigation, login actions, generic website visits, or utility queries should be filtered out.

Return JSON only.

\medskip
\textbf{User Prompt}

Input: A single user behavioral signal

\{USER\_BEHAVIOR\}

\medskip
\textbf{Label Definition}

Assign $y \in \{0,1\}$:

\begin{itemize}
  \item \textbf{1 (Keep)}: The behavior reflects a meaningful informational interest, topic, event, entity, or issue that could correspond to reusable news interests.
  \item \textbf{0 (Filter)}: The behavior is navigational, transactional, login-related, platform-related, generic browsing, random strings, or utility-oriented (e.g., ``google'', ``youtube'', ``facebook login'', ``bank account login'', ``weather today'', URL-only visits).
\end{itemize}

\medskip
\textbf{Examples}

\textbf{Keep (1):}
\begin{itemize}
  \item ``Ukraine Russia conflict latest updates''
  \item ``NVIDIA earnings report Q4''
  \item ``electric vehicle battery technology''
  \item ``US Federal Reserve interest rate decision''
\end{itemize}

\textbf{Filter (0):}
\begin{itemize}
  \item ``google''
  \item ``youtube''
  \item ``bank login''
  \item ``facebook sign in''
  \item ``www.amazon.com''
  \item ``weather''
\end{itemize}

\medskip
\textbf{Output Format}

Return a JSON object only:

\begin{verbatim}
{
  "label": 0 or 1
}
\end{verbatim}

\end{tcolorbox}

\begin{table}[h]
\centering
\setlength{\tabcolsep}{3pt}
\renewcommand{\arraystretch}{1.0}
\small
\begin{tabular}{lcc}
\toprule
\textbf{Reward Setting} & \textbf{Recall@10} & \textbf{NDCG@10} \\
\midrule
Full rewards &0.271  &0.241  \\
\midrule
w/o $R_{\text{align}}$ &0.162  &0.141  \\
w/o $R_{\text{cov}}$ &0.115  &0.093  \\
w/o $R_{\text{spec}}$ &0.196  &0.184  \\
w/o $R_{\text{div}}$ &0.179  &0.144  \\
w/o $R_{\text{struct}}$ &0.193  &0.184  \\
\bottomrule
\end{tabular}
\caption{Reward-level ablation study of the multi-reward GRPO objective.}
\label{tab:reward_ablation}
\end{table}

\section{Reward Decomposition.}
We perform reward-level ablation by removing each component individually. As shown in Table~\ref{tab:reward_ablation}, all rewards are necessary but serve distinct roles. Removing retrieval alignment ($R_{\text{align}}$) significantly degrades performance, as the model generates overly fine-grained interests that are semantically plausible but weakly supported by the news index. Excluding interest coverage ($R_{\text{cov}}$) causes the largest drop: without anchoring queries to major user themes, the model exploits other rewards and drifts toward disconnected interests, leading to severe reward hacking. Removing query specificity ($R_{\text{spec}}$) produces overly coarse and generic interests, weakening personalization and ranking quality. Intra-list diversity ($R_{\text{div}}$) prevents redundancy, while structural validity ($R_{\text{struct}}$) ensures stable downstream execution.

\section{Inference Efficiency Comparison}
\label{app:efficiency}
\begin{table}[t]
\centering
\setlength{\tabcolsep}{3pt}
\renewcommand{\arraystretch}{1.0}
\small
\begin{tabular}{lcc}
\toprule
\textbf{Model} & \textbf{Users/sec} \\
\midrule
Teacher (Qwen2.5-32B) & 2 \\
Student (Qwen2.5-0.5B) & 67 \\
\bottomrule
\end{tabular}
\caption{Inference efficiency comparison using vLLM on a single NVIDIA A100 GPU.}
\label{tab:efficiency}
\end{table}

Table~\ref{tab:efficiency} compares the inference efficiency of the RL-trained teacher model and the distilled student model ($N$=1) under identical hardware settings using vLLM on a single NVIDIA A100 GPU. The teacher model (Qwen2.5-32B) achieves only 2 users per second due to its large parameter size and expensive decoding cost. In contrast, the distilled student model (Qwen2.5-0.5B) reaches 67 users per second, representing over a 30× improvement in throughput. This significant reduction in computational overhead makes it more suitable for the low-latency and high-concurrency requirements of large-scale online recommendation systems.

\end{document}